\documentclass{article}
\usepackage{style}

\usepackage{danudefs}
\usepackage{times}
\usepackage{soul}
\usepackage{url}
\usepackage[hidelinks]{hyperref}
\usepackage[utf8]{inputenc}
\usepackage[small]{caption}
\usepackage{graphicx}
\usepackage{booktabs}
\usepackage{hyperref}
\usepackage{footmisc}
\usepackage{cite}
\usepackage{amsmath, amsfonts, amssymb, amsthm, mathtools}
\usepackage{dsfont}
\usepackage{algorithm, algorithmic}
\usepackage[dvipsnames]{xcolor}
\usepackage{xcolor}
\usepackage{svg}
\title{Assessment of contextualised representations in detecting outcome phrases in clinical trials}

 \author{
 Micheal Abaho$^{1*}$\and
 Danushka Bollegala$^1$\and
 Paula Williamson$^2$\And
 Susanna Dodd$^2$\\
 \affiliations
 {\normalsize $^1$
 Department of Computer Science, University of Liverpool, 
 $^2$
 Department of Health DataScience, University of Liverpool, Liverpoo1, UK
 }\\
 \emails
 {\normalsize
 \{m.abaho,danushka,prw,shinds\}@liverpool.ac.uk}
 }

\begin{document}
\maketitle

\begin{abstract}
\textbf{Background}: Automating the recognition of outcomes  reported  in  clinical  trials  using  machine learning  has  a  huge  potential  of  speeding  up  access  to  evidence  necessary  in  healthcare  decision making. Prior research has however acknowledged inadequate training corpora as a challenge for the Outcome  detection  (OD)  task. Additionally, several  contextualised  representations (embeddings)  like BERT and ELMO have achieved unparalleled success in detecting various diseases, genes, proteins and chemicals,  however,  the same cannot be emphatically stated for outcomes, because these representation models have been relatively under-tested and studied for the OD task.

\textbf{Methods}: We introduce ``EBM-COMET'', a dataset in which 300 Randomised Clinical Trial (RCT) PubMed abstracts are expertly annotated for clinical outcomes. Unlike prior related datasets that use  arbitrary  outcome  classifications,  we  use  labels from a taxonomy recently published to standardise outcome classifications. To extract outcomes, we fine-tune a variety of pre-trained contextualised representations, additionally, we use frozen contextualised and context-independent representations in our custom neural model augmented with clinically informed  Part-Of-Speech embeddings and a cost-sensitive loss function. We adopt strict evaluation for the trained models by rewarding them for correctly identifying full outcome phrases rather than words within the entities i.e. given an outcome phrase “systolic blood pressure”, the models are rewarded a classification score only when they predict all 3 words in sequence, otherwise, they are not 
rewarded. 

\textbf{Results and Conclusion}: We  observe  our  best  model  (BioBERT) achieve  81.5\%  F1,  81.3\%  sensitivity  and 98.0\% specificity. We reach a consensus on which contextualised representations are best suited for detecting outcome phrases from clinical trial abstracts. Furthermore, our best model outperforms scores published on the original EBM-NLP dataset leader-board scores.
\end{abstract}

\section{Introduction}

There is growing recognition of the potential benefits of using readily available sources of clinical information to support clinical research~\cite{Bartlett2019}. Of particular importance is the identification of information about outcomes measured on patients, for example, blood pressure, fatigue, etc. The ability to automatically detect outcome phrases contained within clinical narrative text will serve to maximise the potential of such sources. For example, hospital or GP letters, or free text fields recorded within electronic health records, may contain valuable clinical information which is not readily accessible or analysable without manual or automated extraction of relevant outcome phrases. Similarly, automated identification of outcomes mentioned in trial registry entries or trial publications could help to facilitate systematic review processes by speeding up outcome data extraction. Furthermore, the benefits of automated outcome recognition will be increased further if it extends to categorisation of outcomes within a relevant classification system such as taxonomy  proposed in~\cite{Dodd2018ADiscovery}.
The potential contribution of Natural Language Processing (NLP) to EBM~\cite{Sackett71} has been limited by the scarcity of publicly available annotated corpora~\cite{Nye2018} and the inconsistency in how outcomes are described in different trials~\cite{Dodd2018ADiscovery, Coiera2017, Demner2007}. Nonetheless, rapid advancement in NLP techniques has accelerated NLP-powered EBM research, enabling tasks such as detecting elements that collectively form the basis of clinical questions including Participants/population (P), Interventions (I), Comparators (C), and Outcomes (O)~\cite{Huang2006EvaluationQuestions}. I and C are often collapsed into just I~\cite{jin2018, kim, Nye2018}.

EBM-NLP corpus~\cite{Nye2018} is the only publicly available corpus that can support individual outcome phrase detection. However, this dataset used an arbitrary selection of outcome classifications despite being aligned to Medical Subject Headings (MESH)\footnote{\url{https://www.nlm.nih.gov/mesh/}}. Moreover, it contains flawed outcome annotations~\cite{abaho2019correcting} such as measurement tools and statistical metrics incorrectly annotated as outcomes and others which we mention in section \ref{sec:ebm-nlp}.

In this work, we are motivated by the outcome taxonomy recently built and published to standardise outcome classifications~\cite{Dodd2018ADiscovery}. We work closely with experts to annotate outcomes with classification drawn from this taxonomy.

Several variations of state-of-the-art (SOTA) CLMs that include BioBERT~\cite{lee2020biobert}, SciBERT~\cite{beltagy2019scibert}, ClinicalBERT~\cite{alsentzer2019} and others have recently emerged to aid clinical NLP tasks. Despite their outstanding performance in multiple clinical NLP tasks such as BNER~\cite{stubbs2015, uzuner2007} and relation extraction~\cite{li_jiao2016}, they have been underutilised for the outcome detection task, mainly because of inadequate corpora~\cite{Nye2018}. Given that, clinical trial abstracts (which report outcomes) are part of the medical text on which these CLMs are pre-trained, we leverage transfer learning (TL) and make full use of them to achieve individual outcome detection. The goal in the outcome detection task is to extract outcome phrases from clinical text. For example, in a sentence,``\textit{Among patients who received sorafenib, the most frequently reported \textbf{\underline{adverse events}}, were grade 1 or 2 events of \textbf{\underline{rash}} (73\%), \textbf{\underline{fatigue}} 67\%, \textbf{\underline{hypertension}} (55\%) and \textbf{\underline{diarrhea}} (51\%)}'', we extract all outcome phrases such as those underlined and in bold font. This enables those searching the literature including patients and policymakers to identify research that addresses the health outcomes of most importance to them~\cite{biggane2018}.
Following previous studies that investigated which embeddings are best suited for clinical-NLP text classification tasks~\cite{mascio2020}, we focus this work on probing for some consensus amongst various SOTA domain-specific CLM embeddings, determining which embeddings are best suited for outcome detection. A summary of our contributions includes,

\begin{enumerate}
\item We introduce a novel outcome dataset, EBM-COMET, in which outcomes within randomised clinical trial (RCT) abstracts are expertly annotated with outcome  classifications drawn from~\cite{Dodd2018ADiscovery}.
\item We assess the performance of domain-specific (clinical) context-dependent representations in comparison to generic context-dependent and context-independent representations for the outcome detection task. 
\item We assess the quality in detecting full mention of outcome phrases in comparison to detection of individual words contained in outcome phrases. Ideally, given an outcome phrase ``systolic blood pressure'', full outcome phrase evaluation strictly rewards models for correctly detecting all 3 words in that sequence (exact match), whereas word-level evaluation rewards models for correctly detecting any single word in phrase. The former is particularly informative for the biomedical domain audience~\cite{leaman2015}.
\item We compare the performance of the CLMs in our experimental setup to the current leader-board performance on extracting PIO elements from the original EBM-NLP dataset~\cite{Nye2018}.
\end{enumerate}

\section{Related Work}
\subsection{Outcome detection}
Outcome detection has previously been simultaneously achieved along with Participant and Intervention detection, where researchers aim to classify sentences (extracted from RCT abstracts) into one of P, I and O labels~\cite{wallace2016, jin2018, kiritchenko2010}. Despite being restrained by shortage of expertly labeled datasets, few attempts to create EBM-oriented datasets have been made. Bryon et al.,~\cite{wallace2016} use distant supervision to annotate sentences within clinical trial articles with PICO elements. Dina et al.,~\cite{demner2006} use an experienced Nurse and a medical student to annotate outcomes by identifying and labelling sentences that best summarise the consequence of an intervention. Similarly, other attempts have precisely partitioned PubMed abstracts into sentences that they label one of P, I, and O~\cite{jin2018, kim}. Given the sentence-level annotation adopted in these datasets, it becomes difficult to use them for tasks that require extraction of individual PICO elements~\cite{kang2019, brockmeier2019improving} such as outcome phrase detection. 
Nye et al.,~\cite{Nye2018} recently released EBM-NLP corpus that they built using a mixture of crowd workers (non-experts) and expert workers (with the non-experts being exceedingly more) to annotate individual spans of P, I, O elements within clinical trial articles. This dataset has however been discovered with annotation flaws~\cite{abaho2019correcting} and uses arbitrary outcome classification labels as discussed in section \ref{sec:ebm-nlp}. Cognizant of the growing body of research to standardise classifications of outcomes, we are motivated to annotate a dataset with outcome types drawn from a standardised taxonomy.

\subsection{Transfer Learning (TL)}
TL is a machine learning (ML) approach that enables usage of a model to achieve a task that it was not initially built and trained for~\cite{sun2019transfer}. Usually, the assumption is that, train and test data for a specific task exists, however this is never the case, therefore, TL allows learning across different task domains i.e. the term pre-trained, implies a model was previously trained on a task different from the current target task. 
Context-dependent embeddings such as context2vec~\cite{melamud2016context2vec}, ELMo~\cite{peters2018deep} and BERT~\cite{devlin2018bert} have emerged and outperformed context-independent embeddings~\cite{mikolov2013efficient, pennington2014glove} in various downstream NLP tasks.

Bert variants, SciBERT~\cite{beltagy2019scibert} and ClinicalBERT~\cite{alsentzer2019} yielded performance improvements in the BNER tasks on the BC5DR dataset~\cite{li_jiao2016, dougan2014ncbi}, text-classification tasks like Relation extraction on the ChemProt~\cite{kringelum2016chemprot} and on PICO extraction. Despite being pre-trained on English biomedical text, BioBERT~\cite{lee2020biobert} outperformed generic BERT model ( pre-trained on Spanish biomedical text) in Pharma-CoNER, a multi-classification task for detecting mentions of chemical names and drugs from Spanish biomedical text~\cite{sun2019transfer}. Recently Qiao et al.,~\cite{jin2019probing} discovered that, in comparison to BioBERT, BioELMo (Biomedical ELMo) better clustered entities of the same type such as, an acronym having multiple meanings or a homonym. For example, unlike BioBERT, BioELMo clearly differentiated between ER referring to “Estrogen Receptor” and ER referring to “Emergency Room” in their work.

\section{Materials and Methods}
We design two setups in our assessment approach, where (1) we fine-tune pre-trained biomedical CLMs on the outcome datasets EBM-COMET (introduced in this paper) and \textrm{EBM-NLP$_{\textbf{rev}}$} (a revised version of the original EBM-NLP \cite{abaho2019correcting}) and (2) we augment a neural model to train frozen biomedical embeddings. The aim is to compare the evaluation performance of fine-tuned, frozen biomedical CLM embeddings, generic CLM embeddings and traditional context-independent embeddings such as word2Vec~\cite{mikolov2013efficient} in the outcome detection task defined below.

\paragraph{Outcome Detection Problem (ODP) Task:}
Given a sentence s of n words, $s = w_1,\ldots,w_n$ within an RCT abstract, outcome detection aims to extract an outcome phrase $b = w_x,\ldots,w_d$ within $s$, where $1 \leq x \leq d \leq n$. In order to extract outcome phrases such as $b$, we label each word using the ``BIO'' tagging scheme~\cite{sang1999representing} where ``B'' denotes the first word of the outcome phrase, ``I'' denotes inside the outcome phrase and ``O'' denotes all non-outcome phrase words. 

\subsection{Data} 
\label{sec:data}
\subsubsection{EBM-COMET}
EBM-COMET is prepared to facilitate outcome detection in EBM. Our annotation scheme adopts a widely acknowledged definition of outcome which is ``a measurement or an observation used to capture and assess the effect of treatment such as assessment of side effects (risk) or effectiveness (benefits)''~\cite{william2017}. Previous EBM dataset construction efforts have lacked a standard classification system to accurately inform their annotation process and instead opted for arbitrary labels such as those terms aligned to MeSH~\cite{Nye2018}. We however leverage an outcome taxonomy recently developed to standardise outcome reporting in electronic databases~\cite{Dodd2018ADiscovery}. The taxonomy authors iteratively reviewed how core outcome sets (COS) studies within the Core Outcome Measures in Effectiveness Trials (COMET) database categorised their outcomes. This review culminated into a taxonomy of 38 outcome domains hierarchically classified into 5 outcome types/core areas. 

\subsubsection*{Data collection}
Using the Entrez API~\cite{sayers2009}, we automatically fetch 300 abstracts from open access PubMed. Our search criteria only retrieve articles of type ``Randomised controlled Trial''. We relied on two domain-experts to review these abstracts and eliminate those reporting outcomes in animals (or non-humans). Each eliminated abstract was replaced by another reporting human outcomes from PubMed.

\subsubsection*{Annotation}
The two experts we work with have sufficient experience in reviewing human health outcomes in clinical trials. Some of their work pertaining to outcomes in clinical trials includes~\cite{william2017, williamson2012developing, kirkham2010impact, dwan2013systematic}. These experts jointly annotate granular outcomes within the gathered abstracts resulting into EBM-COMET using guidelines below. We are aware of annotation tools such as BRAT~\cite{brat}, however because of the nature of the annotations i.e. some with contiguous outcome spans, the experts prefer to directly annotate them in Microsoft text documents.   

\subsubsection*{Annotation guidelines}
The annotators are tasked to identify and verify outcome spans and then assign each an outcome domain referenced from the taxonomy partially presented in table \ref{tab:taxonomy} and full presented in Appendix C. The annotators are instructed to assign each span all relevant outcome domains.

\subsubsection*{Annotation heuristics}
\begin{figure*}[h]
\centering
\includegraphics[width=16.5cm, height=6.5cm]{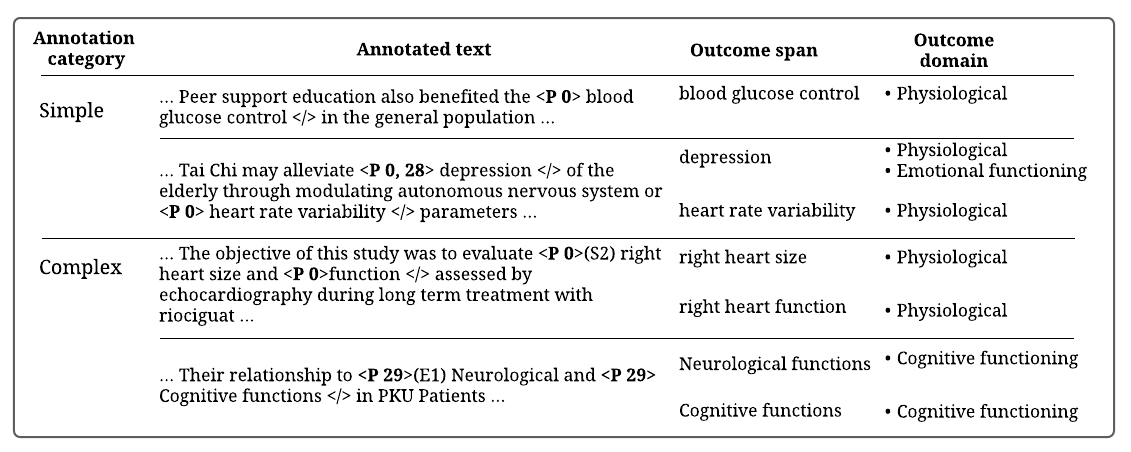}
\caption{Sample annotations of outcomes depicting the annotation style with each example showing the outcome span and its assigned outcome domain label.}
\label{fig:annotation_examples}
\end{figure*}

For annotation purposes, we firstly assign a unique symbol to each outcome domain (domain symbol column in table \ref{tab:taxonomy}). The annotators are then instructed to use these symbols to label the outcome spans they identify. Annotation using these symbols rather than the long domain names is less tedious. Furthermore, we instruct annotators to use xml tags to demarcate the spans, such that an identified span is enclosed within an opening tag with the assigned domain symbol and a closing tag. We refer to easily identifiable outcome spans as simple annotations, and the more difficult ones requiring more demarcation indicators as complex annotations. Figure \ref{fig:annotation_examples} shows examples of the annotations described below, 
\begin{enumerate}
    \item Simple annotations
    \begin{enumerate}
        \item $<$P XX$>$\ldots$<$/$>$: Indicates an outcome belongs to domain XX (where XX can be located in the taxonomy~\ref{tab:taxonomy}).  
        \item $<$P XX, YY$>$\ldots$<$/$>$: Indicates an outcome belongs to both domain XX and YY. 
    \end{enumerate}
    \item Complex annotations \\
    Some spans are contiguous in such a way that, they share a word or words with other spans. For example, two outcomes can easily be annotated as a single outcome because they are conjoined by a dependency word or punctuation such as ``and'', ``or'' and commas. We are however fully aware, that this contiguity previously resulted in multiple outcomes annotated as a single outcome in previous datasets~\cite{abaho2019correcting}. Therefore, annotators are asked to distinctively annotate them as below,
    \begin{enumerate}
        \item Contiguous spans sharing bordering term/s appearing at the start of an outcome span should be annotated as follows, 
        \\
        $<$P XX$>$(S\#)\ldots$<$P XX$>$\ldots$<$/$>$: which indicates that, two outcomes are belonging to domain XX that share \# of words at the start of the annotated outcome span. 
        \item Contiguous spans sharing bordering term/s appearing at the end of an outcome span, should be annotated as follows,
        \\
        $<$P XX$>$(E\#)\ldots$<$P XX$>$\ldots$<$/$>$: The opposite of the notation above indicating that, two outcomes are belonging to domain XX that share \# of words at the end of the annotated outcome span.
    \end{enumerate}
\end{enumerate}

\subsubsection*{Annotation consistency and quality}
In the last phase of the annotation process, the annotations are extracted into a structured format (excel sheet) for the annotators to review them, make necessary alterations based on their expertise judgement as well as handle minor errors (such as wrong opening or closing braces) that result from the manual annotation processes. We do not report inter-annotator agreement because the two annotators did not conduct the process independently, but rather jointly. Having previously worked together on similar annotation tasks, they hardly disagreed but whenever either was uncertain or disagreed, they discussed between themselves and concluded. 

The word, outcome phrase distribution and other statistics of the EBM-COMET are summarised in table \ref{tab:dataset_statistics} with the experimental dataset statistics.

\begin{table}[h]
\begin{tabular}{@{}llc@{}}
\toprule
\textbf{Core area}                                                & \textbf{Outcome domain}                                                    & \textbf{\begin{tabular}[c]{@{}l@{}}Domain\\ Symbol\end{tabular}} \\ \midrule
\begin{tabular}[c]{@{}l@{}}Physiological/Clinical\end{tabular} & \begin{tabular}[c]{@{}l@{}}Physiological/Clinical\end{tabular}          & P 0                                                              \\ \midrule
Death                                                             & Mortality/survival                                                         & P 1                                                              \\ \midrule
Life Impact                                                       & Physical functioning                                                       & P 25                                                             \\
                                                                  & Social functioning                                                         & P 26                                                             \\
                                                                  & Role functioning                                                           & P 27                                                             \\
                                                                  & \begin{tabular}[c]{@{}l@{}}Emotional function\\ ing/wellbeing\end{tabular} & P 28                                                             \\
                                                                  & Cognitive functioning                                                      & P 29                                                             \\
                                                                  & Global quality of life                                                     & P 30                                                             \\
                                                                  & Perceived health status                                                    & P 31                                                             \\
                                                                  & Delivery of care                                                           & P 32                                                             \\
                                                                  & Personal circumstances                                                     & P 33                                                             \\ \midrule
Resource use                                                      & Economic                                                                   & P 34                                                             \\ 
                                                                  & Hospital                                                                   & P 35                                                             \\
                                                                  & \begin{tabular}[c]{@{}l@{}}Need for further \\ intervention\end{tabular}   & P 36                                                             \\
                                                                  & Societal/carer burden                                                      & P 37                                                             \\ \midrule
\begin{tabular}[c]{@{}l@{}}Adverse events\end{tabular}          & \begin{tabular}[c]{@{}l@{}}Adverse events/effects\end{tabular}          & P 38                                                             \\ \bottomrule
\end{tabular}
\caption{A partial version of the taxonomy of outcome classifications developed and used by {[}1{]} to classify clinical outcomes extracted from biomedical articles  published in COMET, Cochrane reviews and clinical trial registry. (Full taxonomy in Appendix C).}
\label{tab:taxonomy}
\end{table}


\subsubsection{\textrm{EBM-NLP$_{\textbf{rev}}$}}
\label{sec:ebm-nlp}
This dataset is a revision of the original hierarchical label's version of EBM-NLP dataset~\cite{Nye2018}. In the hierarchical labels version, the annotated outcome spans were assigned specific labels that include Physical, Pain, Mental, Mortality and Adverse effects. Abaho et al.,~\cite{abaho2019correcting} built \textrm{EBM-NLP$_{\textbf{rev}}$} using a semi-automatic approach that involved POS-tagging and rule-based chunking to correct flaws discovered (by domain-experts) in EBM-NLP. In the evaluation of this revision, classification of outcomes resulted in a significant increase in the F1-score (for all labels) from what it was when using the original EBM-NLP. Some of the major flaws they corrected include, 
\begin{itemize}
    \item Statistical metrics and measurement tools annotated as part of clinical outcomes e.g. \textit{``\textcolor{red}{mean} arterial blood-pressure''} instead of \textit{``arterial blood-pressure''}, \textit{``Quality of life \textcolor{red}{Questionnaire}''} instead of \textit{``Quality of life''}, \textit{``Work-related stress \textcolor{red}{scores}''} \textbf{instead} of \textit{``Work-related stress''}.
    \item Multiple outcomes annotated as a single outcome e.g. \textit{``cardiovascular events-(myocardial infarction, stroke and cardiovascular death)''} instead of \textit{``myocardial infarction''}, \textit{``stroke''}, and \textit{``cardiovascular death''}. 
    \item Inaccurate outcome type annotations e.g. \textit{``Nausea and Vomiting''} labeled as a Mortality outcome instead of a Physical outcome.
    \item Combining annotations in non-human studies with those in human-studies particularly studies reporting outcomes in treating beef cattle. 
\end{itemize}

\subsection{Biomedical Contextual Language Models}
We leverage the datasets to investigate the ODP task performance of 6 different biomedical CLMs (table \ref{tab:clms}) derived from 3 main architectures. 
1) \textbf{BERT}~\cite{devlin2018bert}, a CLM built by learning deep bidirectional representations of input words by jointly incorporating left and right context in all its layers. It works by masking a portion of the input words and thereby predicting missing words in each sentence. BERT encodes a word by incorporating information about words around it within a given input sentence using a self-attention mechanism~\cite{vaswani2017attention} 2) \textbf{ELMo}~\cite{jin2019probing} is a CLM that learns deep bidirectional representations of input words by jointly maximizing the probability of forward and backward directions in a sentence, and 3) \textbf{FLAIR}~\cite{akbik2018contextual}, a character-level bidirectional LM which learns representations of each character by incorporating character information around it within a sequence of words.

\begin{table}[h]
\centering
\resizebox{\columnwidth}{!}{%
\begin{tabular}{@{}lll@{}}
\toprule
\textbf{Model} & \textbf{\begin{tabular}[c]{@{}l@{}}Biomedical\\ Variant\end{tabular}}             & \textbf{Pre-trained on}                                                                                                                                                                                          \\ \midrule
Bert           & \textbf{BioBERT}~\cite{lee2020biobert}                                                          & \begin{tabular}[c]{@{}l@{}}4.5B words from PubMed \\ abstracts + 13.5B words \\ from PubMed Central (PMC) \\ articles.\end{tabular}                                                                                 \\ \cmidrule(lr{1em}){2-3}
              & \textbf{SciBERT}~\cite{beltagy2019scibert}                                                         & \begin{tabular}[c]{@{}l@{}}1.14M Semantic scholar \\ papers~\cite{ammar2018} (18\% \\from Computer science and \\82\% from biomedical\\ domains).\end{tabular}                                                           \\
              \cmidrule(lr{1em}){2-3}
              & \textbf{ClinicalBERT}~\cite{alsentzer2019}                                                    & \begin{tabular}[c]{@{}l@{}}2 million notes in the \\ MIMIC-III v1.4 database~\cite{johnson2016} \\ (hospital care data recorded \\ by nurses). (Bio+Clinical\\BERT is BioBERT pre-trained\\ on the above notes)\end{tabular} \\
              \cmidrule(lr{1em}){2-3}
              & \textbf{\begin{tabular}[c]{@{}l@{}}DischargeSumm\\ aryBERT~\cite{alsentzer2019}\end{tabular}} & \begin{tabular}[c]{@{}l@{}}Similar to ClinicalBERT but \\ only discharge summaries are \\ used (Bio+DischargeSummary\\ BERT is BioBERT pre-trained\\  on the summaries)\end{tabular}                             \\ 
              \midrule
ELMo           & \textbf{BioELMo}~\cite{jin2019probing}                                                         & \begin{tabular}[c]{@{}l@{}}10M PubMed abstracts\\  (ca. 2.64B tokens)\end{tabular}                                                                                                                               \\ \midrule
FLAIR          & \textbf{BioFLAIR}~\cite{sharma2019bioflair}                                                        & 1.8m PubMed abstracts.                                                                                                                                                                                           \\ \bottomrule
\end{tabular}
}
\caption{A catalogue of CLMs used for the outcome detection task }
\label{tab:clms}
\end{table}

\begin{figure*}[h]
\centering
\includegraphics[width=17cm, height=6.0cm]{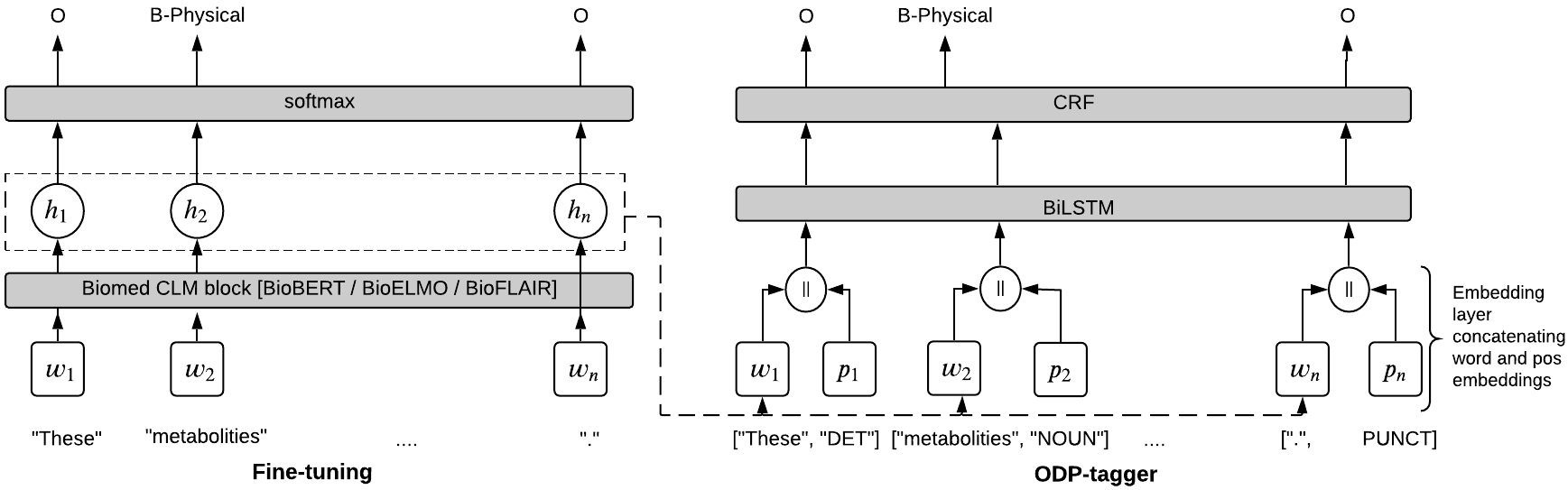}
\caption{BNER for token-level outcome phrase detection, for two setups, left: Fine-tuning and right Feature extraction using ODP-tagger}
\label{fig:fine_tuning_feature_extraction}
\end{figure*}

We begin by further training the pre-trained CLMs in table \ref{tab:clms} in a fine-tuning approach~\cite{howard2018universal},
where the CLMs learn to (1) encode each word $w_i$ into a hidden state $\vec{h}_i$ and (2) predict the correct label given $\vec{h}_i$. Similar to Sun et al..~\cite{sun2019transfer}, we introduce a non-linear softmax layer to predict a label for each $\vec{h}_i$ corresponding to word $w_i$, as shown in figure \ref{fig:fine_tuning_feature_extraction},
where $\vec{h}_i = \mathrm{CLM}(w_i)$, \{BERT-variants, BioELMo, BioFLAIR\} $\in \mathrm{CLM}$. (see Appendix A.1 (Fine-tuning) for more details).

\subsection{ODP-tagger} 
We build ODP-tagger to not only assess context-independent (W2V) representations, but also assess the performance of frozen context-dependent representations for the ODP task. Demonstrated by the dotted line from Fine-tuning to input tokens in figure \ref{fig:fine_tuning_feature_extraction}, is a feature extraction~\cite{peters2019tune} approach, where the tagger’s embedding layer takes as input, a sequence of tokens (sentence) and a sequence of POS terms corresponding to the tokens. We add a POS feature for each token to enrich the model in a manner similar to how prior neural classifiers are enhanced with character and n-gram features~\cite{liu2019neuralclassifier}. Each word/token is therefore represented by concatenating either a pre-trained CLM or a W2V embedding $\vec{w}$ and a randomly initialised embedding for the  corresponding POS term $\vec{p}$. The token embeddings are then encoded to obtain hidden-states for each sequence position,
\begin{align}
    \vec{h}_i = \alpha(\mat{W}[\vec{w}_i;\vec{p}_i]+ b)
\end{align}

where $\vec{w}_i \in \mat{E}^w$ and $\vec{p}_i \in \mat{E}^p$, $\{\mat{E}^w,\mat{E}^p\} \in \R^{n \times d}$ denote Word and POS matrices, each containing d-dimensional embeddings for $n$ words and $n$ corresponding POS terms. $\vec{w}_i$ and $\vec{p}_i$ are the word and POS embeddings representing the $i^{th}$ word and its POS term, $;$ implies a concatenation operation and then $\alpha$ is a linear activation function that generates hidden states for the input words. We then use a condition random field (CRF) layer 
for classification given the hidden state $\vec{h}_i$. A CRF is an undirected graphical model which defines a  conditional probability distribution over possible labels~\cite{lafferty2001conditional}.

All the models are each trained to maximize the probability of the labels given each word $w_i \in s$.
\begin{align}
    \underset{\theta}\argmax P(y_i|\vec{w}_i;\theta)
\end{align}
The training loss objective.
\begin{align}
    loss = - \beta \underset{(S,L)\in\cT}\sum \sum_i^np(y_i|\vec{w}_i)
\end{align}
 where $\beta$ is a scaling factor that empirically sets each labels weights to be inversely proportional to the square root of the label frequency i.e. $\beta = \frac{1}{\sqrt{N_y}}$ and $N_y$ is the number of training samples with ground-truth label $y$. $\cT$ is the training set containing sentences, $\vec{w}_i \in S$ and $y \in L$. 

\begin{table*}[h]
\centering
\resizebox{15cm}{!}{
\begin{tabular}{lcc|lcc}
\toprule
\multicolumn{3}{c}{Fine-tuning}                                                                 &  \multicolumn{3}{c}{Feature extraction}                                                                                \\ \midrule
Model                                                           & \textrm{EBM-NLP$_{\textbf{rev}}$}  & EBM-COMET     & \multicolumn{1}{c}{Model}                                                             & \textrm{EBM-NLP$_{\textbf{rev}}$}  & EBM-COMET     \\ \midrule
W2V                                                             & \_            & \_            & ODP-tagger + W2V                                                                      & 44.0          & 59.3          \\
BERT                                                            & 51.8          & 75.5          & +BERT                                                                                 & 43.2          & 64.2          \\
ELMO                                                            & 49.6          & 71.4          & +ELMO                                                                                 & 43.0          & 61.2          \\
BioBERT                                                         & \textbf{53.1} & \textbf{81.5} & +BioBERT                                                                              & \textbf{48.5} & 69.3          \\
BioELMO                                                         & 52.0          & 75.0          & +BioELMO                                                                              & 46.5          & 62.9          \\
BioFLAIR                                                        & 51.4          & 76.7          & +BioFLAIR                                                                             & 40.7          & 60.5          \\
SciBERT                                                         & 52.8          & 77.6          & +SciBERT                                                                              & 48.1          & \textbf{70.4} \\
ClinicalBERT                                                    & 51.0          & 68.5          & +ClinicalBERT                                                                         & 45.2          & 65.7          \\
Bio+ClinicalBERT                                                & 51.0          & 68.3          & +Bio+ClinicalBERT                                                                     & 45.8          & 66.3          \\
\begin{tabular}[c]{@{}l@{}}Bio+Disc Summary\\ BERT\end{tabular} & 51.0          & 70.0          & \begin{tabular}[c]{@{}l@{}}+Bio+Disc Summary\\                      BERT\end{tabular} & 46.1          & 68.4          \\ \bottomrule
\end{tabular}
}
\caption{Macro-average F1 scores obtained from generic CLMs and their respective In-domain (biomedical) versions for both fine-tuning and ODP-tagger (feature extraction) for token-level detection of outcome phrases from both datasets.}
\label{tab:main_results}
\end{table*}

\subsection{Training}
All models are evaluated on the two datasets discussed in section \ref{sec:data}. These datasets are each partitioned as follows, 75\% for training (train), 15\% for development (dev) and 10\% for testing (test). We exploit the large size of \textrm{EBM-NLP$_{\textbf{rev}}$} (as shown in table \ref{tab:dataset_statistics}) and use its dev set to tune hyperparameters for the ODP-tagger and fine-tuned models (Parameter settings in Appendix B). Each model is trained on a train split of a particular dataset and evaluated on the corresponding test split culminating into results shown in table \ref{tab:main_results}. We use a simple powerful NLP python framework called flair\footnote{\url{https://github.com/flairNLP/flair}} to extract word embeddings from all the BERT and FLAIR variants, and AllenAI\footnote{\url{https://github.com/allenai/bilm-tf}} for BioELMO. Dimensions of the extracted BioFLAIR and BioELMO embeddings are very large, i.e. 7672 and 3072 respectively, which would most likely overwhelm our memory and power-constrained devices during training. Therefore, we apply Principal component Analysis (PCA) dimensionality reduction technique to reduce their dimensions to half their original sizes  while preserving semantic information~\cite{raunak2019effective}. Alongside these embeddings, we evaluate context-independent embeddings which we obtain by training word2vec (W2V) embedding algorithm~\cite{mikolov2013efficient} on 5.5B tokens of PubMed and PMC abstracts. Python and PyTorch~\cite{paszke2019pytorch} deep learning framework are used for implementation, which together with the datasets are made publicly available here \url{https://github.com/MichealAbaho/ODP-tagger}.

\begin{table}[h]
\centering
\resizebox{\columnwidth}{!}{%
\begin{tabular}{@{}l|c|c@{}}
\toprule
                                                                                                      & EBM-COMET          & \textrm{EBM-NLP$_{\textbf{rev}}$}             \\ \midrule
\# of sentences                                                                                        & 5193               & 40092               \\ \midrule
\begin{tabular}[c]{@{}l@{}}\# of train/dev/test\\  sentences\end{tabular}                              & 3895 / 779 / 519   & 30069 / 6014 / 4009 \\ \midrule
\# of outcome labels                                                                                   & 5                  & 6                   \\ \midrule
\begin{tabular}[c]{@{}l@{}}\# of sentences with\\ outcome phrases in \\ train/dev/test\end{tabular}    & 1569 / 451 / 221   & 12481 / 4116 / 3257 \\ \midrule
\begin{tabular}[c]{@{}l@{}}Avg \# of tokens per \\ train/dev/test sentence\end{tabular}                & 20.6 / 21.5 / 21.2 & 25.5 / 26.4 / 25.6  \\ \midrule
\begin{tabular}[c]{@{}l@{}}Avg \# of outcome\\ phrases  per sentence\\  in train/dev/test\end{tabular} & 0.69 / 0.78 / 0.71 & 0.44 / 0.38 / 0.45  \\ \bottomrule
\end{tabular}
}
\caption{Statistics summary of experimental datasets splits. Figures pertaining to Train, Dev and Test sets are separated by a forward slash accordingly. 
}
\label{tab:dataset_statistics}
\end{table}

\subsection{Evaluation results}
Results shown in table \ref{tab:main_results} firstly reveal the superiority of fine-tuning the CLMs in comparison to the ODP-tagger. The best performance across both set-ups is obtained when BioBERT is fine-tuned on the EBM-COMET dataset. However, we observe SciBERT outperform it in the ODP-tagger set-up on the EBM-COMET dataset. Secondly, we observe CLM embeddings produce stronger performances in comparison to context-independent (W2V) embeddings especially with the EBM-COMET dataset. BioFLAIR and ClinicalBERT were the least performing models. For BioFLAIR, we hypothesize that, (1) pre-training on a relatively smaller corpus, (2) it being of much less depth (1-layered BiLSTM) compared to multi-layered BERT and ELMo and (3) downsizing its embeddings using PCA dimensionality reduction are reasons that led to its low performance. For ClinicalBERT, we attributed its struggles to the nature of the corpora on which it is pre-trained. Unlike BioBERT, SciBERT and BioELMo which are pre-trained on PubMed text which is mostly clinical trial abstracts that more often report health outcomes, ClinicalBERT is pre-trained on clinical notes associated with patient hospital admissions \cite{johnson2016}. An additional insight we drew was, performance on the  \textrm{EBM-NLP$_{\textbf{rev}}$} dataset is lower compared to that achieved on EBM-COMET. This was attributed to the annotation inconsistencies in the original EBM-NLP, some of which were resolved in \cite{abaho2019correcting}. Another aspect we closely observed was the runtime. Using a TITAN RTX 24GB GPU, the average runtime for the fine-tuning experiments on EBM-COMET and \textrm{EBM-NLP$_{\textbf{rev}}$} respectively was 7 and 12 hrs. 
On the other-hand, feature extraction (ODP-tagger) experiments were much longer consuming 20 and 36 hours respectively on the same datasets.
Overall, we recommend fine-tuning as a preferred approach for outcome detection, more saw using BioBERT and SciBERT as ideal embedding models.

\begin{table*}[!htb]
\centering
\resizebox{16.5cm}{!}{
\begin{tabular}{@{}lllll@{}}
\toprule
\textbf{Method}                                                            &                                                           & \textbf{Abstract sentence}                                                                                                                                                                                                                     & \multicolumn{2}{c}{\textbf{Full outcome phrase}}                                                                                                                                                                       \\ \midrule
                                                                          & \begin{tabular}[c]{@{}l@{}}Input\\ sentence\end{tabular}  & \textit{\begin{tabular}[c]{@{}l@{}}Among patients who received sorafenib, the most \\ frequently reported \underline{\textbf{adverse events}} were grade 1 or 2 \\ events of \textbf\underline{{rash}} (73\%), \textbf{\underline{fatigue}} (67\%), \textbf{\underline{hypertension}} \\ (55\%) and \textbf{\underline{diarrhea}} (51\%).\end{tabular}} & \textit{\textbf{\begin{tabular}[c]{@{}l@{}}- adverse events\\ - rash\end{tabular}}}                     & \textit{\textbf{\begin{tabular}[c]{@{}l@{}}- fatigue\\ - hypertension\\ - diarrhea\end{tabular}}}            \\ \cmidrule(lr{1em}){2-5}
\begin{tabular}[c]{@{}l@{}}BioBERT+\\ EBM-COMET\end{tabular}               & Output                                                    & \textit{\begin{tabular}[c]{@{}l@{}}Among patients who received sorafenib, the most\\ frequently reported \textcolor[HTML]{0563c1}{\textbf{adverse events}} were grade 1 or 2\\ events of \textcolor[HTML]{0563c1}{\textbf{rash}} (73\%), \textcolor[HTML]{0563c1}{\textbf{fatigue}} (67\%), \textcolor[HTML]{0563c1}{\textbf{hypertension}} \\ (55\%) and \textcolor[HTML]{0563c1}{\textbf{diarrhea}} (51\%).\end{tabular}}   & {\color[HTML]{0563c1} \textit{\textbf{\begin{tabular}[c]{@{}l@{}}- adverse events\\ - rash\end{tabular}}}}     & {\color[HTML]{0563c1} \textit{\textbf{\begin{tabular}[c]{@{}l@{}}- fatigue\\ - hypertension\\ - diarrhea\end{tabular}}}} \\ \cmidrule(lr{1em}){2-5}
\begin{tabular}[c]{@{}l@{}}ODP-tagger+\\ SciBERT\\ +EMB-COMET\end{tabular} & Output                                                    & \textit{\begin{tabular}[c]{@{}l@{}}Among patients who received sorafenib, the most\\ frequently reported adverse events were grade 1 or 2\\ events of \textcolor[HTML]{c00000}{\textbf{rash}} (73\%), \textcolor[HTML]{0563c1}{\textbf{fatigue}} (67\%), \textcolor[HTML]{0563c1}{\textbf{hypertension}} \\ (55\%) and \textcolor[HTML]{0563c1}{\textbf{diarrhea}} (51\%)..\end{tabular}}  & {\color[HTML]{0563c1} \textit{\textbf{\begin{tabular}[c]{@{}l@{}}- fatigue\\ - diarrhea\end{tabular}}}} & {\color[HTML]{0563c1} \textit{\textbf{- hypertension}}}                                                      \\ \midrule
                                                                          & \begin{tabular}[c]{@{}l@{}}Input \\ sentence\end{tabular} & \textit{\begin{tabular}[c]{@{}l@{}}The average \underline{\textbf{duration of operating procedure}} was\\ 1 hour and 35 minutes.\end{tabular}}                                                                                                                      & \multicolumn{2}{l}{\textit{\textbf{- duration of operating procedure}}}                                                                                                                                                \\ \cmidrule(lr{1em}){2-5}
\begin{tabular}[c]{@{}l@{}}BioBERT+\\ EBM-COMET\end{tabular}               & Output                                                    & \textit{\begin{tabular}[c]{@{}l@{}}The average \textcolor[HTML]{c00000}{\textbf{duration of}} \textcolor[HTML]{0563c1}{\textbf{operating procedure}} was\\ 1 hour and 35 minutes.\end{tabular}}                                                                                                                      & \multicolumn{2}{l}{}                                                                                                                                                                                                   \\ \cmidrule(lr{1em}){2-5}
\begin{tabular}[c]{@{}l@{}}ODP-tagger+\\ SciBERT\\ +EMB-COMET\end{tabular} & Output                                                    & \textit{\begin{tabular}[c]{@{}l@{}}The average \textcolor[HTML]{c00000}{\textbf{duration of}} \textcolor[HTML]{0563c1}{\textbf{operating}} \textcolor[HTML]{c00000}{\textbf{procedure}} was \\ 1 hour and 35 minutes.\end{tabular}}                                                                                                                     & \multicolumn{2}{l}{}                                                                                                                                                                                                   \\ \midrule
                                                                          & \begin{tabular}[c]{@{}l@{}}Input \\ sentence\end{tabular} & \textit{\begin{tabular}[c]{@{}l@{}}The objective of this study was to evaluate \\ \underline{\textbf{right heart size}} and \underline{\textbf{function}} assessed by \\ echocardiography during long term treatment with\\  riociguat.\end{tabular}}                                    & \multicolumn{2}{l}{\textit{\textbf{\begin{tabular}[c]{@{}l@{}}- right heart size\\ - right heart function\end{tabular}}}}                                                                                              \\ \cmidrule(lr{1em}){2-5}
\begin{tabular}[c]{@{}l@{}}BioBERT+\\ EBM-COMET\end{tabular}               & Output                                                    & \textit{\begin{tabular}[c]{@{}l@{}}The objective of this study was to evaluate\\ \textcolor[HTML]{0563c1}{\textbf{right heart size}} \textcolor[HTML]{c00000}{\textbf{and function}} assessed by\\ echocardiography during long term treatment with \\ riociguat.\end{tabular}}                                      & \multicolumn{2}{l}{{\color[HTML]{0563c1} \textit{\textbf{- right heart size}}}}                                                                                                                                        \\ \cmidrule(lr{1em}){2-5}
\begin{tabular}[c]{@{}l@{}}ODPtagger+\\ SciBERT+\\ EMB-COMET\end{tabular}  & Output                                                    & \textit{\begin{tabular}[c]{@{}l@{}}The objective of this study was to evaluate\\ \textcolor[HTML]{c00000}{\textbf{right}} \textcolor[HTML]{0563c1}{\textbf{heart size}} \textcolor[HTML]{c00000}{\textbf{and function}} assessed by \\ echocardiography during long term treatment with \\ riociguat.\end{tabular}}                                        & \multicolumn{2}{l}{}                                                                                                                                                                                                   \\ \bottomrule
\end{tabular}
}
\caption{Example outcome detection outputs from best fine-tuned BioBERT and ODP-tagger+SciBERT models.
}
\label{tab:prediction_examples}
\end{table*}

\subsection{Full outcome phrase detection}
Motivated by the need to detect accurate fine-grained information in the medical domain \cite{van2021clinical}, we examine the extent to which our models detect precise mentions of full outcome phrases. To achieve this, we investigate how well the best performing models (Fine-tuned+BioBERT+EBM-COMET and Fine-tuned+BioBERT+EBM-NLP$_{\textbf{rev}}$ from  \autoref{tab:main_results}) can detect full mentions of outcome phrases or otherwise exact matches of outcome phrases in prediction results. We use a strict criteria to evaluate full mention of outcomes, where a classification error FN (False Negative) accounts for the number of full outcome phrases the model fails to detect, which includes partially correctly detected phrases i.e. some of their tokens were misclassified. In table \ref{tab:full_outcome_phrase}, we observe the F1 of the best models drop from 53.1 to 52.4 for \textrm{EBM-NLP$_{\textbf{rev}}$} and 81.5 to 69.6 for EBM-COMET.
This implies that the model struggles to identify full outcome phrases, especially with the \textrm{EBM-NLP$_{\textbf{rev}}$} dataset. Specificity on the other hand is very high for both datasets simply because it is calculated as a True Negative Rate (TNR),
in which case True Negatives (non-outcomes) are certainly so many because they are precisely individual words and therefore are counted word by word as opposed to True positives (actual outcome phrases) that can consist of multiple words. 

\begin{table}[b!]
\centering
\begin{tabular}{@{}lcccc@{}}
\toprule
          & P    & R    & S    & F \\ \midrule
\textrm{EBM-NLP$_{\textbf{rev}}$}   & 53.7 & 51.2 & 99.2 & 52.4 \\
EBM-COMET & 60.8 & 81.3 & 98.0 & 69.6\\ \bottomrule
\end{tabular}
\caption{Precision (P), Recall/Sensitivity (R), Specificity (S) and F1 of outcome entities in \textrm{EBM-NLP$_{\textbf{rev}}$} and EBM-COMET.}
\label{tab:full_outcome_phrase}
\end{table}

We further investigate the errors from the best performing models BioBERT+EBM-COMET (Fine-tuned) and ODP-tagger+SciBERT+EBM-COMET.  In table \ref{tab:prediction_examples}, we show examples of outputs of both models for the ODP task given an input sentence with known actual outcome phrases (underlined).  
Fine-tuned model correctly detects (blue-coded) all full outcome phrase in the first example sentence i.e. Precision (P), Recall/Sensitivity (R) are 100\%, whereas tagger only detects 3/4 outcomes, hence P is 100\%, R is 75\%. Neither of the models
correctly capture full mention of the outcome phrase in the second example, they incorrectly predict some words (red-coded) to not belong to the outcome phrase. While traditionally, results of fine-tuned model would be a P of 100\% and R of 50\% for correct prediction of 2/4 tokens, in our strict full name evaluation, P and R are 0\%, because some tokens in the full outcome phrase are mis-classified in both models i.e. True positives = 0. 
Similarly, in the third example, fine-tuned model achieves P of 100\% and R of 60\% for correct prediction of 3/5 tokens in the traditional evaluation, whereas for the strict full name evaluation, R is 50\% because only 1/2 full outcome phrases are detected. 
We attribute these errors to the length of some outcome phrases with some containing extremely common words such as prepositions (\textit{``of''}). Additionally, we note that the contiguous outcome span annotations (containing several outcomes sharing terms e.g. ``right heart size and function'' in the third example) are rare.

\subsection{Evaluation on the original EBM-NLP}
We additionally fine-tune our best model for the task of detection of all PIO elements in the original EBM-NLP dataset. To be consistent with the original EBM-NLP paper, we consider the token-level detection of the PIO elements task in their work, comparing their evaluation results for hierarchical labels with those we obtain by fine-tuning our best model. Using their published training (4670) and test (190) sets of the starting spans, 
we see fine-tuned BioBERT model outperform the current leader board results~\footnote{\url{https://ebm-nlp.herokuapp.com/}} and the SOTA results published by Brockmeier et al~\cite{brockmeier2019improving} (table \ref{tab:original_ebm_nlp_evaluation}). We attribute this improvement to the fact, unlike the LSTM-CRF and Logreg models in previous SOTA scores, BioBERT's has an internal capability to encode information using self-attention mechanisms to generate context-sensitive representations of words. 

\begin{table}[h]
\centering
\begin{tabular}{@{}lccc@{}}
\toprule
                                                                                             & P             & I             & O             \\ \midrule
Logreg                                                                                       & 45.0          & 25.0          & 38.0          \\
Lstm-crf                                                                                     & 40.0          & 50.0          & 48.0          \\
Brockmeier et.al~\cite{brockmeier2019improving}                                                                    & 70.0          & 56.0          & 70.0          \\
Fine-tuned BioBERT                                                                           & \textbf{71.6} & \textbf{69.0} & \textbf{73.1} \\
\begin{tabular}[c]{@{}l@{}}Fine-tuned BioBERT – Full \\ outcome phrase mentions\end{tabular} & 61.6          & 64.0          & 53.1          \\ \bottomrule
\end{tabular}
\caption{F1 scores of token level detection of PIO elements reported for EBM-NLP hierarchical labels dataset by the EBM-NLP~\cite{Nye2018} leader board,}
\label{tab:original_ebm_nlp_evaluation}
\end{table}

\subsection{Outcome phrase length}
\begin{figure}[h]
\centering
\includegraphics[width=6cm]{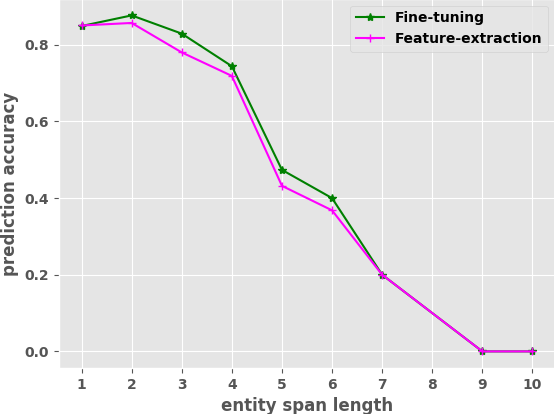}
\caption{Prediction accuracy per entity text-span length.}
\label{fig:entity_span_length}
\end{figure}

To further understand our results, we investigated how well the best models BioBERT+EBM-COMET (Fine-tuned) and ODP-tagger+SciBERT+EBM-COMET (Feature-extraction) detected outcome phrases of varying lengths. We calculate a prediction accuracy as number of correctly predicted outcome-phrases of length x/number of all outcome-phrases of length x, where x ranged from 1-10. As observed in figure \ref{fig:entity_span_length},  the fine-tuned model slightly outperforms the ODP-tagger especially for outcome phrases having 3-6 words (i.e. 3-6 entity span length). However, it is also clear that both models struggled to accurately detect outcome phrases containing 7 or more words.

\section{Conclusion}
In this work, we present EBM-COMET, a dataset of clinical trial abstracts with outcome annotations to facilitate EBM tasks. Experiments showed that CLMs perform much better on EBM-COMET than they do on EBM-NLP, indicating it is suited for ODP task especially because it is well aligned to standardised outcome classifications. Our assessment showed) fine-tuned models consistently outperform and converge faster than feature extraction, particularly pre-trained BioBERT and SciBERT embedding models. Additionally, we show the significance of accurate detection of full mention of granular outcome phrases which is beneficial for clinicians searching for this information.

\bibliographystyle{vancouver}
\bibliography{assessment_clms}

\section*{Appendix}
\appendix

\typeout{IJCAI-19 Multiple authors example}



\renewcommand*{\thesection}{\Alph{section}}
\renewcommand*{\thesubsection}{\arabic{subsection}}


\section{Adapting CLMs to Outcome Detection Task}
\subsection{Fine-tuning}
The biomedical CLMs presented under section 3.2 are fine-tuned for the Outcome Detection (ODP) task. Given an input sentence containing $n$ words/tokens, e.g. $s = w_1,\ldots,w_n$, the CLMS are used to encode each a word $w_i$ to obtain a hidden state representation $\vec{h}_i = {CLM}(w_i)$, where $1 \leq i \leq n$,  \{BERT-variants, BioELMo, BioFLAIR\}$\in {CLM}$ and $\vec{h}_i \in \R^{n \times d}$ (i.e. $\vec{h}_i$ is a vector of size $d$). We then apply softmax function to return a probability of each label for each position in the sentence $s$, $y = \mathrm{softmax}(\mat{W}\cdot\vec{h}_i + b)$, where $\mat{W} \in \R^{|\cL| \times k}$ i.e. $\mat{W}$ is a matrix with dimensions $|\cL|$ (size of label set) $\times k$ (hidden-state size). $\cL$ represents the set of outcome type target labels. Given the probability distribution the softmax generates at each position, we use $\underset{\theta}\argmax P(y|\vec{w}_n;\theta)$ to to return the predicted outcome type label. 

\subsection{Building an Outcome Detection Model (ODP-tagger)}
In this work, we augment a BiLSTM model with in-domain resources including medically oriented part-of-speech tags (POS) and PubMed word2vec vectors~\cite{mikolov2013efficient}. We then train the model on \textrm{EBM-NLP$_{\textbf{rev}}$} incorporating a class distribution balancing factor which essentially aims to regularize the multiway softmax loss with a balanced weighting across multiple classes. The conscious effort of augmenting a regular BiLSTM was indeed re-enumerated with a visible gradual improvement in dev set F1 scores for the ODP task as table \ref{tab:odp_results} presents. Below sections cover the augmentation steps.

\subsubsection{Custom trained biomedical POS}
We compare the performance of 3 Part-Of-Speech (POS) taggers, which include, 2 popular generic and fully established Natural language Processing (NLP) libraries, spaCy\footnote{\url{https://spacy.io/}}~\cite{Bartlett2019}, Stanford Core NLP\footnote{\url{https://nlp.stanford.edu/software/tagger.html}}~\cite{manning2014stanford}, and a tagger specifically tuned for POS tagging tasks on biomedical text (Genia-Tagger)~\cite{tsuruoka2005developing}. The Genia-Tagger is pre-trained on a collection of articles extracted from the MEDLINE database~\cite{kim2003genia}. To avoid any biased analysis in the comparative study, spaCy and Stanford Core NLP are also customised for biomedical text by training them on a corpus of 6,700 Medline sentences (MedPOST) annotated with 60 POS tags~\cite{smith2004medpost}. These 3 taggers are each used to provide POS features to input samples (words) for a task to classify outcome phrases into five outcome types that include Physical, Pain, Mental, Mortality, Adverse effects and Other as predefined in \textrm{EBM-NLP$_{\textbf{rev}}$} dataset. A BiLSTM network and a softmax classification layer are used to complete this task. The model using trained Stanford tagger outperforms the other two models (table \ref{tab:pos_table}), and as a result, we use Stanford Core NLP for POS tagging in the proceeding ODP task.

\begin{table}[!htb]
\centering
\begin{tabular}{@{}lc@{}}
\toprule
                        & \textrm{EBM-NLP$_{\textbf{rev}}$} (F1\%)     \\ \midrule
BiLSTM-spaCY-MedPOST    & 80.5 \\
BiLSTM-stanford-MedPOST & 81.3 \\
BiLSTM-Genia-Tagger     & 79.0 \\ \bottomrule
\end{tabular}
\caption{Macro-average F1 scores in a text classification task of Outcomes in EBM-NLP$_{\textbf{rev}}$ corpus. Biomedical POS taggers including spaCY-MedPOST, stanford-MedPOST and Genia-Tagger are used to provide POS features which alongside the text are used in training the BiLSTM model.}
\label{tab:pos_table}
\end{table}

\subsubsection{Context-Independent PubMed word2vec vectors (W2V) }
We train word2vec (W2V) on 5.5B tokens of PubMed and PMC abstracts to obtain these vectors. These fixed vectors are later replaced by the pre-trained CLMs in the feature extraction approach during evaluation.

\subsubsection{Probing for a loss function for the ODP-tagger}
We assess 3 cost-sensitive functions premised on a log-likelihood objective $\log p(y|w)$, (log probability of label $y$ given input word $w$) to identify a suitable learning loss for the ODP-tagger experiments. 
\begin{align}
	\label{eq:odp_loss}
    \mathrm{ODP_{loss}} = - \underset{(S,L)\in\cT}\sum \sum_i^np(y_i|\vec{w}_i)
\end{align}
where $\cT$ is the training set containing sentences, $\vec{w}_i \in S$ and $y \in L$.

\subsubsection{Imputed Inverse loss (IIL) function}
Empirically setting each labels’ weights to be inversely proportional to the label frequency.  A relatively simple heuristic that has been widely adopted~\cite{wang2017learning}. 
\begin{align}
	\label{eq:iil_loss}
    \mathrm{IIL} = {\beta} \cdot \mathrm{ODP_{loss}}
\end{align}
We check two variants of the scaling factor $\beta$ in the Imputed Inverse Loss equation $\mathrm{IIL}_1$, $\beta = \frac{1}{N_y}$ and a smoothed version $\mathrm{IIL}_2$, $\beta = \frac{1}{\sqrt{N_y}}$, where $N_y$ is the number of training samples labelled $y$ or frequency of ground truth label $y$.

\subsubsection{Class balanced loss (CB)}
The Class balanced loss proposed by Cui et al.,~\cite{cui2019class} discusses the concept of effective number of samples to capture the diminishing marginal benefits of incrementing the samples of a class. Due to the intrinsic similarities among real-world data, increasing the sample size of a class might not necessarily improve model-performance.  Cui et al.,~\cite{cui2019class} introduces a weighting factor that is inversely proportional to the effective number samples $E_n$.

Where $E_n = \frac{1-\beta}{1-\beta^{n_y}}$, $\beta=\frac{N-1}{N}$, $N$ is dataset size and $n_y$ is the sample size of label $y$, $\beta^{n_y} = \frac{n_y-1}{n_y}$.
\begin{align}
	\label{eq:iil_loss}
    \mathrm{CL} = \frac{1}{E_n}\mathrm{ODP_{loss}}
\end{align}

\subsubsection{Focal loss (FL)}
Focal loss assigns higher weights to harder examples and lower ones to the easier examples~\cite{lin2017focal}.  It introduces a scaling factor $(1 - p)^\lambda$. $\lambda$ is a focusing parameter in the loss function which decays to zero as the confidence in the correct class increases hence automatically down weighting the contribution of easy examples in the training and rapidly focusing on harder examples.

\begin{align}
	\label{eq:iil_loss}
    \mathrm{FL} = -\alpha_y(1 - P_y)^\lambda\mathrm{ODP_{loss}}
\end{align}

where $\alpha$ is a weighting factor, $\alpha\in[0,1]$, $\alpha_y$  is set to $\frac{1}{N_y}$ , $N_y$ is the number of training samples for class $y$, $P_y$ is the probability of ground truth label $y$. We do not hypertune the focusing parameter $\lambda$, and instead set it to $\lambda = 2$ based on having achieved good results in examples~\cite{cui2019class}.

\begin{table}[hbt!]
\centering
\begin{tabular}{@{}lc@{}}
\toprule
                                                                                      & \textrm{EBM-NLP$_{\textbf{rev}}$}     \\ \midrule
BiLSTM                                                                                & 27.0 \\
BiLSTM + $\mathrm{IIL}_1$ & 37.0 \\
BiLSTM + $\mathrm{IIL}_2$                          & 38.0 \\
BiLSTM + CB                                                                           & 37.0 \\
BiLSTM + FL                                                                           & 19.0 \\ \bottomrule
\end{tabular}
\caption{F1 \% scores in the ODP task for various cost-sensitive loss functions on the EBM-NLP$_{\textbf{rev}}$ corpus. BiLSTM$^{*}$ implies the model was training with default ODP$_{loss}$ objective as shown in \eqref{eq:odp_loss}}
\label{tab:build_odp}
\end{table}

Results in table \ref{tab:build_odp} indicate both $\mathrm{IIL}$ variants and CB are quite competitive, however we chose $\mathrm{IIL}_2$ particularly because it slightly outperforms all the other tested  $\mathrm{IIL}_2$ for the objective loss function.

\subsubsection{Introducing an undersampling hyper-parameter (US)}
\label{sec:ebm-nlp}
In this strategy, we randomly undersample the majority class of the dataset by a specified percentage. 
The objective of the ODP-tagger is to minimize the Imputed Inverse loss (IIL$_2$) derived from the preceding section which probes for a suitable loss function,  

\begin{align}
	\label{eq:iil_loss}
    \mathrm{IIL}_2 = - \frac{1}{\sqrt{N_y}}\underset{(S,L)\in\cT}\sum \sum_i^np(y_i|\vec{w}_i)
\end{align}

\begin{table}[b!]
\centering
\resizebox{\columnwidth}{!}{
\begin{tabular}{@{}llc@{}}
\toprule
  & Model                                      & F1  \\ \midrule
1 & BiLSTM                                     & 32.5  \\
2 & BiLSTM + POS                               & 37.9  \\
3 & BiLSTM + POS + W2V                         & 41.1  \\
4 & BiLSTM + POS + W2V + $\mathrm{IIL}$                 & 43.2  \\
5 & BiLSTM + POS + W2V + $\mathrm{IIL}$ + $\mathrm{US}_{50}$       & 43.6  \\
6 & BiLSTM + POS + W2V + $\mathrm{IIL}$ + $\mathrm{US}_{50}$ + CRF & 44.0  \\ 
\midrule
7 & BiLSTM + POS$_{St}$  + W2V$_{Pb}$ + $\mathrm{IIL}_2$                 & 
42.8 (1.5)  \\
8 & BiLSTM + POS$_{St}$  + W2V$_{Pb}$ + $\mathrm{IIL}_2$ + $\mathrm{US}_{50}$       & 43.2 (1.9)  \\
9 & BiLSTM + POS$_{St}$  + W2V$_{Pb}$ + $\mathrm{IIL}_2$ + $\mathrm{US}_{50}$ + CRF & 44.3 (1.4)  \\ \bottomrule
\end{tabular}
}
\caption{F1 \% scores in the ODP task resulting from incrementally augmenting the BiLSTM with various components to build the ODP-tagger. BiLSTM$^{*}$ implies the model was training with default ODP$_{loss}$ objective as shown in \eqref{eq:odp_loss}, POS$_{St}$ denotes POS tagging by Stanford CoreNLP tagger, W2V$_{Pb}$ denotes Word2Vec trained using PubMed articles (Only non-contextual embeddings are tested in this investigation because they have smaller dimensions), $\mathrm{IIL}_2$ denotes Imputed Inverse loss, $\mathrm{US}_{50}$ denotes Undersampling majority class by 50\%. Exps 1-5 use a softmax classifier which is replaced by a CRF in 5. Exps 7-9 report the mean and (standard deviation) over 5 random train/test splits}
\label{tab:odp_results}
\end{table}

Table \ref{tab:odp_results} results are emblematic of the positive impact each of the different strategies had in architecting the ODP-tagger. We observe slight performance improvements upon adopting $\mathrm{US}_{50}$  (a strategy in which the majority class is undersampled by 50\% during training) and replacement of the softmax with a CRF for classification.  We observe cumulative gains in performance of 5.4\%, 3.2\% and 2.1\%  upon adding POS$_{St}$, W2V$_{Pb}$ and $\mathrm{IIL}_2$ respectively. On the otherhand, adopting $\mathrm{US}_{50}$ and replacement of the softmax with a CRF for classification lead to slight improvements of 0.4\% each.

We are aware that the improvements narrated above can dramatically change given new splits of the data, particularly the slight improvements brought about by $\mathrm{US}_{50}$ and the CRF. Therefore, to account for this, we check for the robustness of the improvements brought about by $\mathrm{US}_{50}$ and the CRF by measuring performance across 5 different randomly split train and test sets.  
The mean and (standard deviation) across the 5 experiments of the random splits are reported in Exps 7, 8 and 9. Results obtained in 8 and 9 show that both $\mathrm{US}_{50}$ and the CRF respectively lead to substantial improvements in performance when added to the ODP-tagger. Later on, we hypertune multiple parameters to obtain the optimal parameter settings (\ref{tab:parameter_settings}) for fine-tuning and feature extraction experiments.


\section{Hyper-parameter Tuning}
The tuned ranges for the hyper-parameters used in our models are included in table \ref{tab:parameter_settings}.

\begin{table}[h!]
\centering
\resizebox{\columnwidth}{!}{
\begin{tabular}{@{}lcc@{}}
\toprule
                 \multicolumn{3}{c}{Fine-tuning}   \\ \midrule
                 & Tuned range                 & Optimal 
                 \\ \midrule
Learning rate    & {[}1e-5,1e-4, 1e-3, 1e-2{]} & 1e-5  \\
Train Batch size & {[}16, 32{]}                & 32    \\
Epochs           & {[}3, 5, 10{]}              & 10    \\
Sampling \% (US) & {[}50, 75, 100{]}           & 100   \\
Optimizer        & {[}Adam, SGD{]}             & Adam  \\ \midrule 
\multicolumn{3}{c}{ODP-tagger} \\ \midrule
Learning rate    & {[}1e-4, 1e-3, 1e-2, 1e-1{]} & 1e-1  \\
Train Batch size & {[}50, 150, 250, 300{]}    & 300    \\
Epochs           & {[}60, 80, 120, 150{]}     & 60     \\
Sampling \% (US) & {[}10, 25, 50, 75{]}       & 50     \\
Optimizer        & {[}Adam, SGD{]}            & SGD    \\ \bottomrule
\end{tabular}
}
\caption{Hyper-parameter tuning details in the feature extraction approach for the fine-tuned CLMs and the ODP-tagger (feature extraction).}
\label{tab:parameter_settings}
\end{table}

\onecolumn
\section{A classification taxonomy of outcome domains suitable for retrieval of outcome phrases from clinical text}

\begin{table}[h!]
\centering
\resizebox{16.5cm}{!}{
\begin{tabular}{@{}llcl@{}}
\toprule
\textbf{Core area} & \textbf{Outcome domain}         & \textbf{Domain symbol} & \textbf{Explanation}                                                                                                                                                                                                                                                                                                                                                                                                                                                                                                                                                                       \\ \midrule
Physiological      & Physiological/Clinical          & P 0                    & \begin{tabular}[c]{@{}l@{}}Includes measures of physiological function, signs and\\ symptoms, laboratory (and other scientific) measures \\ relating to physiology.\end{tabular}                                                                                                                                                                                                                                                                                                                                                                                                           \\ \midrule
Death              & Mortality/survival              & P 1                    & \begin{tabular}[c]{@{}l@{}}Includes overall (all-cause) survival/mortality and \\ cause-specific survival/mortality, as well as composite\\ survival outcomes that include death (e.g. disease-free \\ survival, progression-free survival, amputation-free survival).\end{tabular}                                                                                                                                                                                                                                                                                                        \\ \midrule
Life impact        & Physical functioning            & P 25                   & \begin{tabular}[c]{@{}l@{}}Impact of disease/condition on physical activities of \\ daily living (for example, ability to walk, independence,\\ self-care, performance status, disability index, motor skills,\\  sexual dysfunction. health behaviour and management).\end{tabular}                                                                                                                                                                                                                                                                                                       \\
                  & Social functioning              & P 26                   & \begin{tabular}[c]{@{}l@{}}Impact of disease/condition on social functioning (e.g.\\ ability to socialise, behaviour within society, communication,\\ companionship, psychosocial development, aggression, \\ recidivism, participation).\end{tabular}                                                                                                                                                                                                                                                                                                                                     \\
                  & Role functioning                & P 27                   & \begin{tabular}[c]{@{}l@{}}Impact of disease/condition on role (e.g. ability to care for\\ children, work status).\end{tabular}                                                                                                                                                                                                                                                                                                                                                                                                                                                            \\
                  & Emotional functioning/wellbeing & P 28                   & \begin{tabular}[c]{@{}l@{}}Impact of disease/condition on emotions or overall wellbeing\\ (e.g. ability to cope, worry, frustration, confidence, perceptions\\ regarding body image and appearance, psychological status, \\ stigma, life satisfaction, meaning and purpose, positive affect,\\ self-esteem, self-perception and self-efficacy).\end{tabular}                                                                                                                                                                                                                              \\
                  & Cognitive functioning           & P 29                   & \begin{tabular}[c]{@{}l@{}}Impact of disease/condition on cognitive function (e.g. memory\\ lapse, lack of concentration, attention); outcomes relating to \\ knowledge, attitudes and beliefs (e.g. learning and applying \\ knowledge, spiritual beliefs, health beliefs/knowledge).\end{tabular}                                                                                                                                                                                                                                                                                        \\
                  & Global quality of life          & P 30                   & \begin{tabular}[c]{@{}l@{}}Includes only implicit composite outcomes measuring global\\ quality of life.\end{tabular}                                                                                                                                                                                                                                                                                                                                                                                                                                                                      \\
                  & Perceived health status         & P 31                   & \begin{tabular}[c]{@{}l@{}}Subjective ratings by the affected individual of their relative\\ level of health.\end{tabular}                                                                                                                                                                                                                                                                                                                                                                                                                                                                 \\
                  &                                 &                        & \begin{tabular}[c]{@{}l@{}}Includes outcomes relating to the delivery of care, including\\ - adherence/compliance, withdrawal from intervention \\ e.g. time to treatment failure). \\ - tolerability/acceptability of intervention.\\ - appropriateness, accessibility, quality and adequacy of \\ intervention.\\ - patient preference, patient/carer satisfaction (emotional \\ rather than financial burden).\\ - process, implementation and service outcomes (e.g.\\ overall health system performance and the impact of service\\ provision on the users of services).\end{tabular} \\
                  & Personal circumstances          & P 33                   & \begin{tabular}[c]{@{}l@{}}Includes outcomes relating to patient’s finances, home \\ and environment.\end{tabular}                                                                                                                                                                                                                                                                                                                                                                                                                                                                         \\ \midrule
Resource use       & Economic                        & P 34                   & \begin{tabular}[c]{@{}l@{}}Includes general outcomes (e.g. cost, resource use) not\\ captured within other specific resource use domains.\end{tabular}                                                                                                                                                                                                                                                                                                                                                                                                                                     \\
                  & Hospital                        & P 35                   & \begin{tabular}[c]{@{}l@{}}Includes outcomes relating to inpatient or day care hospital\\ care (e.g. duration of hospital stays, admission to ICU).\end{tabular}                                                                                                                                                                                                                                                                                                                                                                                                                           \\
                  & Need for further intervention   & P 36                   & \begin{tabular}[c]{@{}l@{}}Includes outcomes relating to,\\ - medication (e.g. concomitant medications, pain relief)\\ - surgery (e.g. caesarean delivery, time to transplantation)\\ - other procedures (e.g. dialysis-free survival, mode of delivery)\end{tabular}                                                                                                                                                                                                                                                                                                                      \\
                  & Societal/carer burden           & P 37                   & \begin{tabular}[c]{@{}l@{}}Includes outcomes relating to financial or time implications\\ on carer or society as a whole e.g. need for home help, entry\\  to institutional care, effect on family income\end{tabular}                                                                                                                                                                                                                                                                                                                                                                     \\ \midrule
Adverse events     & Adverse events/effects          & P 38                   & \begin{tabular}[c]{@{}l@{}}Includes outcomes broadly labelled as some form of unintended\\ consequence of the intervention e.g. adverse events/effects,\\ adverse reactions, safety, harm, negative effects, toxicity, \\ complications, sequelae. Specifically named adverse events\\ should be classified within the appropriate taxonomy domain \\ above\end{tabular}                                                                                                                                                                                                                   \\ \bottomrule
\end{tabular}
}
\caption{A taxonomy of outcome classifications developed and used by~\cite{Dodd2018ADiscovery} to classify clinical outcomes extracted from biomedical articles published in repositories that include Core Outcome Measures in Effectiveness Trials (COMET), Cochrane reviews and clinical trial registry}
\label{tab:full_taxonomy}
\end{table}

\end{document}